\documentclass{article}

\usepackage{microtype}
\usepackage{graphicx}
\usepackage{booktabs} 
\usepackage{subcaption}
\usepackage{xcolor}
\usepackage[ruled,vlined]{algorithm2e}
\usepackage{pgfplots}
\pgfplotsset{compat=1.17}
\usepackage{tikz}
\usepackage[export]{adjustbox}
\usepackage{tablefootnote}

\usepackage{hyperref}
\usepackage[normalem]{ulem}
\usepackage{enumitem}

\usepackage[accepted]{icml2022}

\usepackage{amsmath}
\usepackage{amssymb}
\usepackage{mathtools}
\usepackage{amsthm}

\theoremstyle{plain}

\theoremstyle{definition}

\theoremstyle{remark}

\usepackage[textsize=tiny]{todonotes}
\usepackage{authblk}

\icmltitlerunning{Patch Gradient Descent (PatchGD)}
\title{Patch Gradient Descent: Training Neural Networks \\ on Very Large Images}
\author[1,2]{Deepak K. Gupta$^*$}
\author[1]{Gowreesh Mago$^*$}
\author[1]{Arnav Chavan$^*$}
\author[2]{Dilip K. Prasad}
\date{}

\affil[1]{Transmute AI Lab, Indian Institute of Technology, ISM Dhanbad, India}
\affil[2]{Dept. of Computer Science, UiT The Arctic University of Norway, Tromso, Norway}
\begin{document}
\twocolumn[
\maketitle





\icmlkeywords{Machine Learning, ICML}

\vskip 0.3in
]



\let\thefootnote\relax\footnotetext{* indicates equal contribution}

\begin{abstract}
Traditional CNN models are trained and tested on relatively low resolution images ($<300$ px), and cannot be directly operated on large-scale images due to compute and memory constraints. We propose Patch Gradient Descent (PatchGD), an effective learning strategy that allows to train the existing CNN architectures on large-scale images in an end-to-end manner. PatchGD is based on the hypothesis that instead of performing gradient-based updates on an entire image at once, it should be possible to achieve a good solution by performing model updates on only small parts of the image at a time, ensuring that the majority of it is covered over the course of iterations. PatchGD thus extensively enjoys better memory and compute efficiency when training models on large scale images. PatchGD is thoroughly evaluated on two datasets - PANDA and UltraMNIST with ResNet50 and MobileNetV2 models under different memory constraints. Our evaluation clearly shows that PatchGD is much more stable and efficient than the standard gradient-descent method in handling large images, and especially when the compute memory is limited. 
\end{abstract}

\section{Introduction}
Convolutional neural networks (CNNs) are considered among the most vital ingredients for the rapid developments in the field of computer vision. This can be attributed to their capability of extracting very complex information far beyond what can be obtained from the standard computer vision methods. For more information, we refer the reader to the recently published comprehensive reviews  \cite{khan2020air, Zewen2021tnn, Alzubaidi2021jbd}. 

\begin{figure}
    \centering
    \includegraphics[scale=0.21]{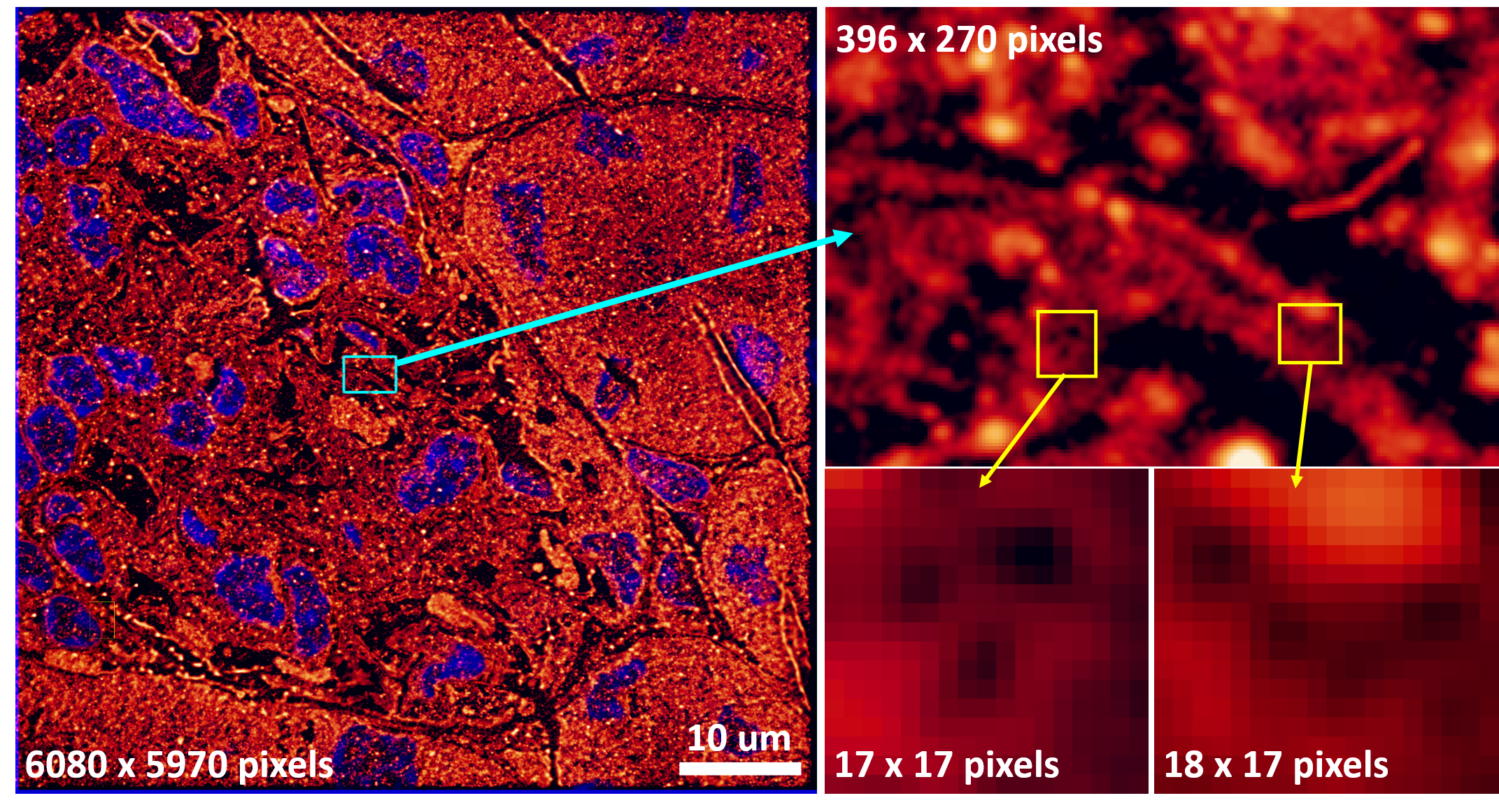}
    \caption{Example nanoscopy image (left) of a mouse kidney cryo-section approximately 1/12th of the area of a single field-of-view of the microscope, chosen to illustrate the level of details at different scales. The bottom right images show that the smallest features in the image of relevance can be as small as a few pixels (here 5-8 pixels for the holes)\cite{villegas2022chip}.}
    \label{fig-nano}
\end{figure}


With the recent technological developments, very large images are obtained from data acquisition in the fields of microscopy \cite{khater2020review,schermelleh2019super}, medical imaging \cite{aggarwal2021npj}, and earth sciences \cite{huang2018agricultural,amani2020google}, among others. Recently, there has been a drive to use deep learning methods in these fields as well. In particular, several deep learning methods have been proposed to handle the images from the microscopy domain \cite{orth2017microscopy,dankovich2021challenges,sekh2020learning,sekh2021physics}, however,  the big data challenge of applying CNNs to analyze such images is immense, as we demonstrate in Figure \ref{fig-nano}. High content nanoscopy involves taking nanoscopy images of several adjacent fields-of-view and stitching them side-by-side to have a full perspective of the biological sample, such as a patient's tissue biopsy, put under the microscope. There is information at multiple scales embedded in these microscopy images \citep{villegas2022chip}, with the smallest scale of features being only a few pixels in size. Indeed, such dimensions of images and levels of details are a challenge for the existing CNNs.

Existing deep learning models using CNNs are predominantly trained and tested on relatively low resolution regime (less than \mbox{$300\times 300$ pixels}). This is partly because the widely used image benchmarking datasets such as ILSVRC(ImageNet dataset) \cite{imagenet2010large} for classification and PASCAL VOC \cite{everingham2009pascal} for object detection/segmentation consist of low-resolution images in a similar range, and most of the existing research has been towards achieving state-of-the-art (SOTA) results on these or similar datasets. Using these models on high-resolution images leads to quadratic growth of the associated activation size, and this in turn leads to massive increase in the training compute as well as the memory footprint. Further, when the available GPU memory is limited, such large images cannot be processed by CNNs.


\begin{figure}
\centering
\begin{tikzpicture}
\begin{axis}[
          width=0.8\linewidth, 
          table/col sep=comma,
          xlabel=Epochs, 
          ylabel=Accuracy (in fraction),
          label style={font=\small},
          tick label style={font=\scriptsize},
          legend style={font=\scriptsize, at={(0.97,0.45)}}, 
          x tick label style={rotate=0} 
        ]
\addplot [mark=none, dashed, blue] table[x=Epoch, y=baseline16GB, col sep=comma]{images/final.csv};
\addlegendentry{Gradient descent (16 GB)};
\addplot [mark=none, solid, thick, blue] table[x=Epoch, y=patchgd16GB, col sep=comma]{images/final.csv};
\addlegendentry{PatchGD (16 GB)};
\addplot [mark=none, dashed, red] table[x=Epoch, y=baseline4GB, col sep=comma]{images/final.csv};
\addlegendentry{Gradient descent (4 GB)};
\addplot [mark=none, solid, thick, red] table[x=Epoch, y=patchgd4GB, col sep=comma]{images/final.csv};
\addlegendentry{PatchGD (4 GB)};
\end{axis}
\end{tikzpicture}\vspace{-1em}
\caption{Performance comparison of standard CNN and PatchGD (ours) for the task of classification of UltraMNIST digits of size $512\times512$ pixels using ResNet50 model. Two different computational memory budgets of 16 GB and 4GB are used, and it is demonstrated that PatchGD is relatively stable for the chosen image size, even for very low memory compute.}
 \label{fig-ultramnist1}
\end{figure}
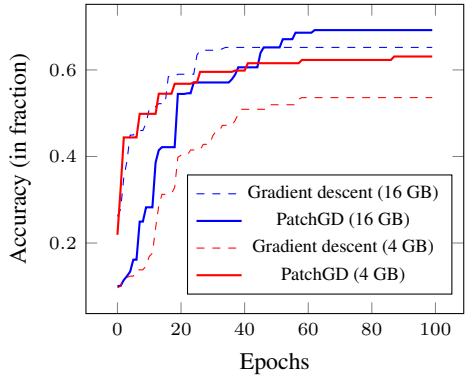

There exist very limited works that address the issue of handling very large images using CNNs. The most common approach among these is to reduce the resolution of the images through downscaling. However, this can lead to a significant loss of information associated with the small-scale features, and it can adversely affect the semantic context associated with the image. An alternate strategy is to divide the image into overlapping or non-overlapping tiles and process the tiles in a sequential manner. However, this approach does not assure that the semantic link across the tiles will be preserved and it can hinder the learning process. Several similar strategies exist that attempt to learn the information contained in the large images, however, their failure to capture the global context limits their use.  

In this paper, we present a novel CNN training pipeline that is capable of handling very large images. We point out here that `large images' should not be plainly interpreted in terms of the number of pixels that they comprise, rather an image should be considered too large to be trained with CNNs if the respective computational memory budget available for it is small. For example, while training a a ResNet50 classification model with images of size $10,000 \times 10,000$ might be hardly possible on a GPU card of 48 GB memory, a GPU memory of 12 GB could be good enough to train the same model on $512\times 512$ size images. Further, when the same $512\times 512$ size images are trained on a GPU memory limit of 4 GB, these might be looked at as too large. 

Figure \ref{fig-ultramnist1} presents a better understanding of the problem outlined above. We consider here the task of classification of the UltraMNIST digits \cite{gupta2022umnist} into one of the 10 predefined classes labelled from 0-9. UltraMNIST images used here comprise 3-5 MNIST digits of extremely varying scale and the sum of the digits ranges between 0-9. The label class of each image corresponds to the sum of the contained digits. More details related to the UltraMNIST classification problem are presented in Appendix \ref{app-umnist}. We consider here images of size $512 \times 512$ pixels and pose the problem to be solved at two different computational memory budgets. We consider the two cases of GPU memory limits of 4 GB and 16 GB. For the base CNN model, we use ResNet50 \cite{he2016deep} architecture and employ the standard training approach. We refer to this approach as Gradient descent (GD). We further present results obtained using the proposed training pipeline, referred as \emph{PatchGD}. Abbreviated for Patch Gradient Descent, it is a scalable training method designed to build neural networks with either very large images, or very low memory compute or a combination of both. 

The efficacy of PatchGD is evident from the results in Figure \mbox{\ref{fig-ultramnist1}} where PatchGD outperforms the conventional GD method for 16 GB as well as 4 GB memory limit. While the difference in performance is 4\% at 16 GB, it grows to a remarkable margin of  13\% difference in the accuracy measure at 4 GB. The classification problem at 4 GB memory compute is intended to replicate the real-world challenges when dealing with large images. With only 4 GB in hand, the image size of $512 \times 512$ is already too large to be used for training a ResNet50 model, and this leads to the inferior performance shown in Figure \ref{fig-ultramnist1}. However, PatchGD is stable even at this low memory regime, and this can be attributed to its design that makes it invariant to image size to a large extent. We describe the details of the method later in the paper as well as demonstrate through experimental results on a variety of image sizes that PatchGD is capable of adapting the existing CNN models to work with very large images even if the available GPU memory is limited.

\textbf{Contributions. }To summarize, the contributions of this paper can be listed as follows.
\begin{itemize}
    \item We present \emph{Patch Gradient Descent (PatchGD)}, a novel strategy to train neural networks on very large images in an end-to-end manner. 
    \item Due to its inherent ability to work with small fractions of a given image, PatchGD is scalable on small GPUs, where training the original full-scale images may not even be possible.
    \item PatchGD reinvents the existing CNN training pipeline in a very simplified manner and this makes it compatible with any existing CNN architecture. Moreover, its simple design allows it to benefit from the pre-training of the standard CNNs on the low-resolution data.
\end{itemize}

\section{Related Work}
This paper aims at improving the capability of CNNs in handling large-scale images in general. To our knowledge there is only very limited research work in this direction and we discuss them in this section.
Most works that exist focus on histopathological datasets since these are popular sources of large images. The majority of existing works employ pixel-level segmentation masks, which are not always available. For example, \citet{Iizuka2020,DBLP:journals/corr/LiuGNDKBVTNCHPS17} perform patch-level classification based on labels created from patchwise segmentation masks available for the whole slide images (WSI), and then feed it to a RNN to obtain the final WSI label. 
\citet{Braatz} use goblet cell segmentation masks to perform patch-level feature extraction. However, these approaches require labelled segmentation data, are computationally expensive, feature learning is very limited, and the error propagation is higher. 

 Another set of methods focus on building a compressed latent representation of the large input images using existing pretrained models or unsupervised learning approaches.
 For example, \citet{Lai2022} use U-Net autoencoder and stack them into a cube, which is then fed to another module to obtain slide-level predictions. \citet{Tellez2018NeuralIC} explore the use of different encoding strategies including reconstruction error minimization, contrastive learning and adversarial feature learning to map high-resolution patches to a lower-dimensional vector. \citet{https://doi.org/10.48550/arxiv.2004.07041} extend this work and use multi-task learning to get better representations of patches than their unsupervised counterparts. 
One important limitation of this class of methods is that the encoding network created from unsupervised learning is not always the strong representative of the target task. 

There exist several methods that use pretrained models derived from other other tasks as feature extractors and the output is then fed to a classifier. Example methods include using Cancer-Texture Network (CAT-Net) and  Google Brain (GB) models as feature extractors \cite{Kosaraju2022}, or additionally using similar datasets for fine-tuning \cite{Brancati2021GigapixelHI}. Although these methods gain advantage from transfer learning, such two-stage decoupled pipelines propagate errors through under-represented features and the performance of the model on the target task is hampered. In this paper, we propose a single step approach that can be trained in an end-to-end manner on the target task.


Several research works have focused on identifying the right patches from the large images and use them in a compute-effective manner to classify the whole image. \citet{Naik2020} propose to construct the latent space using randomly selected tiles, however, this approach does not preserve the semantic coherence across the tiles and fails to extract features that are spread across multiple tiles. 
\citet{Campanella2019} consider this as a multi-instance learning approach, assigning labels to top-K probability patches for classification. \citet{9178453,Huang2022} propose a patch-based training, but make use of streaming convolution networks. \citet{Sharma2021ClustertoConquerAF} cluster similar patches and performs cluster-aware sampling to perform WSI and patch classification. \citet{DBLP:journals/corr/abs-2104-03059} use a patch scoring mechanism and patch aggregator network for final prediction, however they perform downsampling for patch scoring which may cause loss of patch-specific feature important for WSI. \citet{DBLP:journals/corr/abs-2102-10212} progressively increases the resolution and localize the regions of interest dropping the rest equivalent to performing hard adaptive attention. \citet{DIPALMA2021102136} train a teacher model at high-resolution and performs knowledge distillation for the same model at lower resolution. \citet{DBLP:journals/corr/abs-1905-03711} perform attention sampling on downsampled image and derive an unbiased estimator for the gradient update. However their method involves downsampling for attention which may loose out some vital information. It is important to note that all such methods which employ patch selection and knowledge distillation are orthogonal to our work and can be easily combined with our work. However, this is beyond the scope of this paper.

With the recent popularity of Transformer-based methods for vision-based tasks, \citet{Chen_2022_CVPR} proposed a self-supervised learning objective for pre-training large-scale vision transformer at varying scale. Their method involves a hierarchical vision transformer which leverages the natural hierarchical structure inherent in WSI. However their method requires a massive pre-training stage which is not always feasible. Also their method is specific to WSI rather than more general image classification and involves training multiple large-scale transformers. Our method on the other hand, targets more general image classification task and does not involve large scale pre-training, rather it directly works over any existing CNN model.

\section{Approach}
\begin{figure*}[!ht]
    \centering
    \begin{subfigure}{0.9\linewidth}
    \centering
    \includegraphics[width=\textwidth]{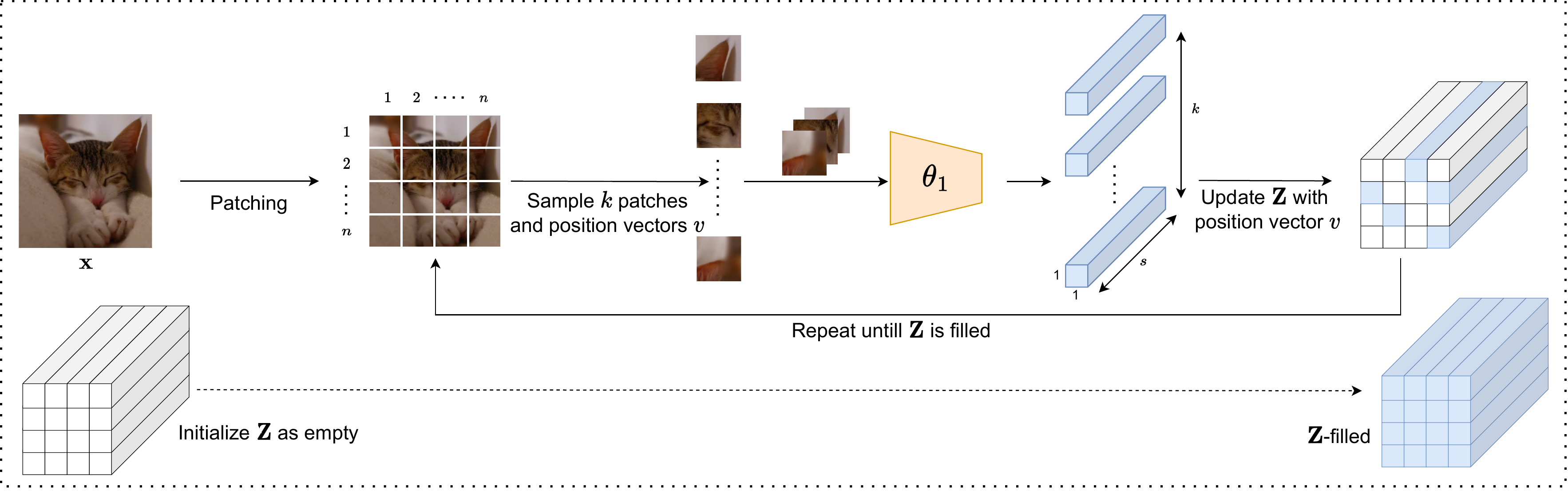}
    \vspace{-1.7em}
    \caption{Pipeline for the filling of $\mathbf{Z}$ block, also referred as $\mathbf{Z}$-filling. }
    \label{fig-zfilling}
    \end{subfigure}\vspace{1em}
        \begin{subfigure}{0.9\linewidth}
        \centering
    \includegraphics[width=\textwidth]{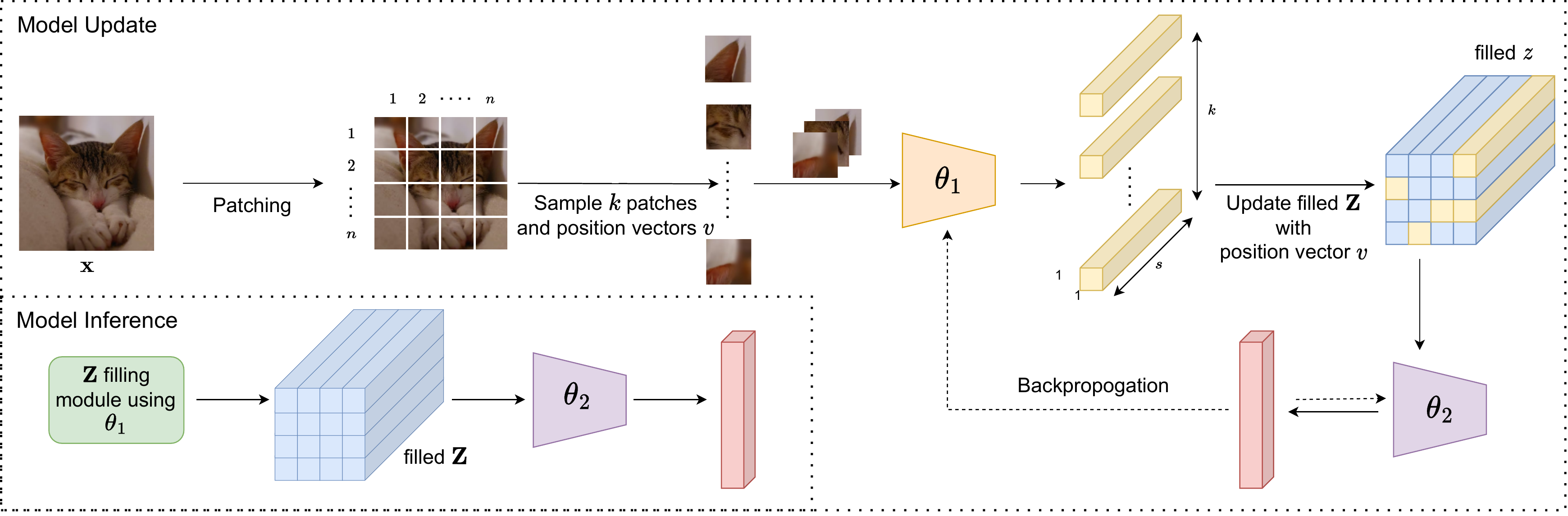}
    \caption{Model update and model inference.}
    \label{fig-modelpipe}
    \end{subfigure}
    \caption{Schematic representations of the pipelines demonstrating working of different components of the PatchGD process.}
    \label{fig-schema}
\end{figure*}

\subsection{General description}
\label{general-desc}
\textit{Patch Gradient Descent (PatchGD)} is a novel CNN training strategy that can train networks with high-resolution images. It is based on the hypothesis that, rather than performing gradient-based updates on an entire image at once, it should be possible to achieve a good solution by performing model updates on only small parts of the image at a time, ensuring that the majority of it is covered over the course of iterations. However, even if only a portion of the image is used, the model is still trainable end-to-end with PatchGD.

Figure \ref{fig-schema} presents a schematic explanation of the PatchGD method. At the core of PatchGD lies the construction or filling of $\mathbf{Z}$ block, a deep latent representation of the full input image. Irrespective of which parts of the input are used to perform model updates, $\mathbf{Z}$ builds an encoding of the full image based on information acquired for different parts of it from the previous few update steps. We further explain the use of the $\mathbf{Z}$ block using the diagram shown in Figure \ref{fig-zfilling}. As can be seen, $\mathbf{Z}$ is primarily an encoding of an input image $\mathbf{X}$ obtained using any given model parameterized with weights $\boldsymbol\theta_1$. The input image is divided into $m \times n$ patches and each patch is processed as an independent image using $\boldsymbol\theta_1$. The size of $\mathbf{Z}$ is always enforced to be $m\times n\times s$, such that patch $\mathbf{x}_{ij}$ in the input space corresponds to the respective $1\times 1\times s$ segment in the $\mathbf{Z}$ block. 

The process of $\mathbf{Z}$-filling spans over multiple steps, where every step involves sampling $k$ patches and their respective positions from $\mathbf{X}$ and passing them as a batch to the model for processing. The output of the model combined with the positions are then used to fill the respective parts of $\mathbf{Z}$. Once all the $m \times n$ patches of $\mathbf{X}$ are sampled, the filled form of $\mathbf{Z}$ is obtained. The concept of filling $\mathbf{Z}$ is employed by PatchGD during model training as well as inference stages. To build an end-to-end CNN model, we add a small subnetwork comprising convolutional and fully-connected layers that processes the information contained in $\mathbf{Z}$ and transforms it into a vector of $c$ probabilities as desired for the task of classification. It is important to note that the cost of adding this small sub-network is generally negligible. The pipelines for model training and inference are shown in Figure \ref{fig-modelpipe}. During training, model components $\boldsymbol\theta_1$ as well as $\boldsymbol\theta_2$ are updated. Based on a fraction of patches sampled from the input image, the respective encodings are computed using the latest state of $\boldsymbol\theta_1$ and the output is used to update the corresponding entries in the already filled $\mathbf{Z}$. The partially updated $\mathbf{Z}$ is then used to further compute the loss function value and the model parameters are updated through back-propagation. 

\subsection{Mathematical formulation}
In this section, we present a detailed mathematical formulation of the proposed PatchGD approach and describe its implementation for the model training and inference steps. For the sake of simplicity, we tailor the discussion towards training of a CNN model for the task of classification.

Let $f_{\boldsymbol\theta}: \mathbb{R}^{M\times N\times C}\rightarrow\mathbb{R}^c$ denote a CNN-based  model parameterized by $\boldsymbol\theta$ that takes an input image $\mathbf{X}$ of spatial size $M \times N$ and $C$ channels, and computes the probability of it to belong to each of the $c$ pre-defined classes. To train this model, the following optimization problem is solved.

\begin{equation}
    \underset{\boldsymbol\theta}{\text{min}} \enskip \mathcal{L}(f(\boldsymbol\theta; \mathbf{X}), \mathbf{y}),
\end{equation}
where $\{\mathbf{X}, \mathbf{y}\} \in \mathcal{D}$ refers to the data samples used to train the network and $\mathcal{L}(\cdot)$ denotes the loss function associated with the training. Traditionally, this problem is solved in deep learning using the popular mini-batch gradient descent approach where updates are performed at every step using only a fraction of the data samples. We present below the formulation of standard gradient descent followed by the formulation our PatchGD method.

\textbf{Gradient Descent (GD). }Gradient descent in deep learning involves performing model updates using the gradients computed for the loss function over one or more image samples. With updates performed over one sample at a time, referred as stochastic gradient descent method, the model update at the $i^{\text{th}}$step can be mathematically stated as
\begin{equation}
    \boldsymbol\theta^{(i)} = \boldsymbol\theta^{(i-1)} - \alpha \frac{\mathrm{d}\mathcal{L}\hfill}{\mathrm{d}\boldsymbol\theta^{(i-1)}},
\end{equation}

where $\alpha$ denotes the learning rate. However, performing model updates over one sample at a time leads to very slow convergence, especially because of the noise induced by the continuously changing descent direction. This issue is alleviated in mini-batch gradient descent method where at every step, the model weights are updated using the average of gradients computed over a batch of samples, denoted here as $\mathcal{S}$. Based on this, the update can be expressed as
\begin{equation}
    \boldsymbol\theta^{(i)} = \boldsymbol\theta^{(i-1)} - \frac{\alpha}{N(\mathcal{S})} \sum_{\mathbf{X}\in\mathcal{S}}\frac{\mathrm{d}\mathcal{L}^{(\mathbf{X})}\hfill}{\mathrm{d}\boldsymbol\theta^{(i-1)}}
    \label{eq-gd2}
\end{equation}
and $N(S)$ here denotes the size of the batch used. 
As can be seen in Eq. \ref{eq-gd2}, if the size of image samples $s \in \mathcal{S}$ is very large, it will lead to large memory requirements for the respective activations, and under limited compute availability, only small values of $N(\mathcal{S})$, sometimes even just 1 fits into the GPU memory. This should clearly demonstrate the limitation of the gradient descent method, when handling large images. This issue is alleviated by our PatchGD approach and we describe it next.

\textbf{PatchGD. }As described in Section \ref{general-desc}, PatchGD avoids model updates on an entire image sample in one go, rather it computes gradients using only part of the image and updates the model parameters. In this regard, the model update step of PatchGD can be stated as
\begin{equation}
    \boldsymbol\theta^{(i, j)} = \boldsymbol\theta^{(i, j-1)} - \frac{\alpha}{k\cdot N(\mathcal{S}_i)} \sum_{\mathbf{X}\in\mathcal{S}_i}\sum_{p\in\mathcal{P}_{\mathbf{X},j}}\frac{\mathrm{d}\mathcal{L}^{(\mathbf{X}, p)}\hfill}{\mathrm{d}\boldsymbol\theta^{(i,j-1)}}.
    \label{eq-pgd1}
\end{equation}
 In the context of deep learning, $i$ here refers to the index of the mini-batch iteration within a certain epoch. Further, $j$ denotes the inner iterations, where at every inner iteration, $k$ patches are sampled from the input image $\mathbf{X}$ (denoted as $\mathcal{P}_{\mathbf{X},j}$) and the gradient-based updates are performed as stated in Eq. \ref{eq-pgd1}. Note that for any iteration $i$, multiple inner iterations are run ensuring that the the majority of samples from the full set of patches that are obtained from the tiling of $\mathbf{X}$ are explored.

 In Eq. \ref{eq-pgd1}, $\boldsymbol\theta^{(i, 0)}$ denotes the initial model to be used to start running the inner iterations on $\mathcal{S}_i$ and is equal to $\boldsymbol\theta^{(i-1, \zeta)}$, the final model state after $\zeta$ inner iterations of patch-level updates using $\mathcal{S}_{i-1}$. For a more detailed understanding of the step-by-step model update process, please see \mbox{Algorithm \ref{alg:model-update}}. As described earlier, PatchGD uses an additional sub-network that looks at the full latent encoding $\mathbf{Z}$ for any input image $\mathbf{X}$. Thus the parameter set $\boldsymbol\theta$ is extended as $\boldsymbol\theta = [\boldsymbol\theta_1, \boldsymbol\theta_2]^{\intercal}$, where the base CNN model is $f_{\boldsymbol\theta_1}$ and the additional sub-network is denoted as $g_{\boldsymbol\theta_2}$.

 Algorithm \ref{alg:model-update} describes model training over one batch of $B$ images, denoted as $\mathcal{X} \in \mathbb{R}^{B \times M \times N \times C}$. As the first step of the model training process, $\mathbf{Z}$ corresponding to each $\mathbf{X} \in \mathcal{X}$ is initialized. The process of filling of $\mathbf{Z}$ is described in Algorithm \ref{alg:z-filling}. For patch $\mathbf{x}_{ab}$, the respective $\mathbf{Z}[a,b,:]$ is updated using the output obtained from $f_{\boldsymbol\theta_1}$. Note here that $\boldsymbol\theta_1$ is loaded from the last state obtained during model update on the previous batch of images. During the filling of $\mathbf{Z}$, no gradients are stored for backpropagation.

Next the model update process is performed over a series of $\zeta$ inner-iterations, where at every step $j \in \{1, 2, \hdots, \zeta\}$, $k$ patches are sampled per image $\mathbf{X}\in \mathcal{X}$ and the respective parts of $\mathbf{Z}$ are updated. Next, the partly updated $\mathbf{Z}$ is processed with the additional sub-network $\boldsymbol\theta_2$ to compute the class probabilities and the corresponding loss value. Based on the computed loss, gradients are backpropagated to perform update of $\boldsymbol\theta_1$ and $\boldsymbol\theta_2$. Note that we control here the frequency of model updates in the inner iterations through an additional term $\epsilon$. Similar to how a batch size of 1 in mini-batch gradient descent introduces noise and adversely affects the convergence process, we have observed that gradient update per inner-iteration leads to sometimes poor convergence. Thus, we introduce gradient accumulation over $\epsilon$ steps and update the model accordingly. 
Note that gradients are allowed to backpropagate only through those parts of $\mathbf{Z}$ that are active at the $j^{\text{th}}$ inner-iteration. During inference phase, $\mathbf{Z}$ is filled using the optimized $f_{\boldsymbol\theta_1^*}$ as stated in Algorithm \ref{alg:z-filling} and then the filled version of $\mathbf{Z}$ is used to compute the class probabilities for input $\mathbf{X}$ using $g_{\boldsymbol\theta_2^*}$.

\begin{algorithm}[tb]
   \caption{Model Training for 1 iteration}
   \label{alg:model-update}
\begin{algorithmic}
   \STATE {\bfseries Input:} Batch of input images $\mathcal{X}\in \mathbb{R}^{B \times M\times N \times C}$, Pre-trained feature extractor $f_{\boldsymbol\theta_1}$, Classifier head $g_{\boldsymbol\theta_2}$, Patch size $p$, Inner iterations $\zeta$, Patches per inner iteration $k$, Batch size $B$, Learning rate $\alpha$, Grad. Acc. steps $\epsilon$
   \STATE \textbf{Initialize:} $\mathbf{Z} = \mathbf{0}^{B \times m\times n \times c}, \mathbf{U}_1 = \mathbf{0}, \mathbf{U}_2 = \mathbf{0}$
   \STATE $\mathbf{Z} \leftarrow \mathbf{Z}$-filling$(\mathbf{X}, f_{\boldsymbol\theta_1}, p) \text{ forall } \mathbf{X} \in \mathcal{X}$
   \STATE $ f_{\boldsymbol\theta_1} \leftarrow \texttt{start\_gradient}(f_{\boldsymbol\theta_1})$
   \FOR{$j : 1$ to $\zeta$} 
   \FOR{$\mathbf{X}$ in $\mathcal{X}$}
   \STATE 
   $\{\mathcal{P}_{\mathbf{X},j}, v\} = \texttt{patch\_sampler}(\mathbf{X}, k), $ \\
   \STATE $\mathcal{P}_{\mathbf{X},j}\in \mathbb{R}^{  p\times p \times C \times k}$  
   \STATE $\mathbf{z} = f_{\boldsymbol\theta_1} 
   (\mathcal{P}_{\mathbf{X},j})$ 
   \STATE $\mathbf{Z}[v] = \mathbf{z} $ // \text{Update the positional embeddings}
   \STATE $\mathbf{y}_{\text{pred}} = g_{\boldsymbol\theta_2}(\mathbf{Z})$
   \STATE $ \mathcal{L} = \texttt{calculate\_loss}(\mathbf{y}, \mathbf{y}_{\text{pred}})$
   \STATE $ \mathbf{U}_1 = \mathbf{U}_1 + \text{d}\mathcal{L}/\text{d}\boldsymbol\theta_1, \mathbf{U}_2 = \mathbf{U}_2 + \text{d}\mathcal{L}/\text{d}\boldsymbol\theta_2$
   \ENDFOR
   \IF{$j \% \epsilon = 0$}
   \STATE $\mathbf{U}_1 = \mathbf{U}_1/\epsilon$, $\mathbf{U}_2 = \mathbf{U}_2/\epsilon$
   \STATE $\boldsymbol\theta_1 = \boldsymbol\theta_1 - \alpha  \mathbf{U}_1$
   \STATE $\boldsymbol\theta_2 = \boldsymbol\theta_2 - \alpha \mathbf{U}_2$
   \STATE $ \mathbf{U}_1 = \mathbf{0}, \mathbf{U}_2 = \mathbf{0}$
   \ENDIF
   \ENDFOR
\end{algorithmic}
\end{algorithm}

\begin{algorithm}[tb]
   \caption{Filling of the $\mathbf{Z}$ block (referred as $\mathbf{Z}$-filling)}
   \label{alg:z-filling}
\begin{algorithmic}
   \STATE {\bfseries Input:} Input image $\mathbf{X}\in \mathbb{R}^{M\times N \times C}$, Pre-trained feature extractor $f_{\boldsymbol\theta_1}$, Patch size $p$, $n = (N/p), m = (M/p)$
   \STATE \textbf{Initialize:} $\mathbf{Z} \in \mathbb{R}^{m\times n \times s}, \boldsymbol\theta_1 = \texttt{stop\_graph}(\boldsymbol\theta_1)$
   \REPEAT 
   \STATE $\mathbf{x}_{a, b} = \texttt{patch\_extractor}(\mathbf{X}, a, b)$
   \STATE $\mathbf{x}_{a, b} \in \mathbb{R}^{ p\times p \times C}$ 
   \STATE $\mathbf{z}_{a, b} = f_{\boldsymbol\theta_1}(\mathbf{x}_{a, b}), \mathbf{z}_{i}\in \mathbb{R}^{ 1\times 1 \times s}$
   \STATE $\mathbf{Z}[a, b] = \mathbf{z}_{a, b}$
   \UNTIL{all patches sampled}
   \STATE \textbf{Return $\mathbf{Z}$}
\end{algorithmic}
\end{algorithm}

\section{Experiments}
We demonstrate here the efficacy of PatchGD through multiple numerical experiments on two benchmark datasets comprising large images with features at multiple scales. 

\subsection{Experimental setup}

\textbf{Datasets. }For the experiments presented in this paper, we consider two datasets: UltraMNIST \cite{gupta2022umnist} and Prostate cANcer graDe Assessment (PANDA) \cite{pandas} datasets. UltraMNIST is a classification dataset and each sample comprises 3-5 MNIST digits of varying scales placed at random locations in the image such that the sum of the digits lies between 0-9. PANDA dataset comprises high-resolution histophathological images, and for this study, we consider a maximum image resolution of $4096 \times 4096$ pixels. Note that unlike the aforementioned approaches, we do not make use of any segementation masks for PANDA. Therefore, the complete task boils down to taking an input high-resolution image and then classifying them into 6 categories based on the International Society of Urological Pathology (ISUP) grade groups.
More details related to the datasets can be found in Appendix \ref{app-datasets}. 

\textbf{CNN models. }We consider two popular CNN architectures: ResNet50 \cite{he2016deep} and MobileNetV2 \cite{sandler2018mobilenetv2}. ResNet50 is a popular network from the residual networks family and forms backbone for several models used in a variety of computer vision tasks (such as object detection and tracking). Thus, we demonstrate the working of PatchGD on primarily this model. MobileNetV2 is a light-weight architecture which is commonly employed for edge-devices, and it would be of interest to see how it performs with large images under limited memory scenarios.

\textbf{Implementation details. } We follow the same hyperparameters across our experiments for a fair comparison. Exact details are stated in Appendix \ref{app-train-hyp}. We report classification accuracy and quadratic weighted kappa (QWK) for PANDA dataset. PyTorch is the choice of framework to implement both the baselines and the PatchGD. We follow 4GB, 16GB and 24GB memory constraints to mimic the popular deep learning GPU memory limits. Latency is calculated on 40GB A100 GPU, completely filling the GPU memory.

\subsection{Results}

\begin{table}[]
\footnotesize
\centering
\caption{Performance scores for standard Gradient Descent and our PatchGD method obtained using ResNet50 architectures on the task of UltraMNIST classification with images of size $512\times512$.}
\resizebox{\columnwidth}{!}{
\begin{tabular}{@{}lccc@{}}
\toprule
Method          & Patch size        & Memory (in GB)        & Accuracy  \\ \midrule
GD              & -                 & 16                    &  65.2            \\
GD-extended     & -               & 16                    &  50.5            \\
PatchGD         & 256               & 16                    & \textbf{69.2}              \\
GD              & -               & 4                     & 53.6              \\
GD-extended     & -               & 4                     & 52.5              \\
PatchGD         & 256               & 4                     &  \textbf{63.1}             \\
\bottomrule
\end{tabular}}
\label{tab-umnist-resnet}
\end{table}
\begin{table}[]
\footnotesize
\centering
\caption{Performance scores for standard Gradient Descent and our PatchGD method on the task of UltraMNIST classification with images of size $512\times512$ obtained using MobileNetV2 architecture.}
\resizebox{\columnwidth}{!}{
\begin{tabular}{@{}lccc@{}}
\toprule
Method          & Patch size        & Memory (in GB)        & Accuracy \%\\ \midrule
GD              & -                 & 16                    & 67.3              \\
GD-extended     & -                 & 16                    & 64.3              \\
PatchGD         & 256               & 16                    & \textbf{83.7}              \\
GD              & -                 & 4                     & 67.7              \\
GD-extended     & -                 & 4                     & 60.0              \\
PatchGD         & 256               & 4                     & \textbf{74.8}             \\
\bottomrule
\end{tabular}}
\label{table-umnist-mbnet}
\end{table}
\textbf{UltraMNIST classification. }The performance of PatchGD for UltraMNIST has already been shown in Figure \ref{fig-ultramnist1}. More detailed results are presented in Tables 
\ref{tab-umnist-resnet} and \ref{table-umnist-mbnet}. For both the architectures, we see that PatchGD outperforms the standard gradient descent method (abbreviated as GD) by large margins. Our approach employs an additional sub-network $g_{\boldsymbol\theta_2}$, and it can be argued that the gains reported in the paper are due to it. For this purpose, we extend the base CNN architectures used in GD and report the respective performance scores in Tables \ref{tab-umnist-resnet} and \ref{table-umnist-mbnet} as GD-extended.

For both the architectures, we see that PatchGD outperforms GD as well as GD-extended by large margins. For ResNet50, the performance difference is even higher when we have a low memory constraint. At 4 GB, while GD seems unstable with a performance dip of more than 11\% compared to the 16 GB case, our PatchGD approach seems to be significantly more stable. For MobileNetV2, the difference between PatchGD and GD is even higher at 16GB case, thereby clearly showing that PatchGD blends well with even light-weight models such as MobileNetV2. For MobileNetV2, we see that going from 16 GB to 4 GB, there is no drop in model performance, which demonstrates that MobileNetV2 can work well with GD even at low memory conditions. Nevertheless, PatchGD still performs significantly better. The underlying reason for this gain can partly be attributed to the fact that since PatchGD facilitates operating with partial images, the activations are small and more images per batch are permitted. We also observe that the performance scores of GD-extended are inferior compared to even GD. ResNet50 and MobilenetV2 are optimized architectures and we speculate that addition of plain convolutional layers in the head of the network is not suited due to which the overall performance is adversely affected.

\begin{table*}[ht]
\centering
\caption{Performance scores obtained using Resnet50 on PANDA dataset for Gradient Descent (GD) and Patch Gradient Descent (PatchGD). In case of 512 image size, 10\% sampling leads to only one patch, hence 30\% patches are chosen.}
\resizebox{\textwidth}{!}{
\begin{tabular}{@{}ccccccccll@{}}
\toprule
Method                  & Resolution & Patch Size & Sampling \% & Mem. Constraint & \# Parameters (M) (G) & Latency (imgs/sec) & Accuracy \% & QWK \\ \midrule
GD                      & 512      & -          & -           & 16                  & 23.52             & 618.05             & 44.4     & 0.558   \\
PatchGD                 & 512      & 128        & 30*         & 16                  & 26.39             & 521.42             & 44.9     & 0.576\\
GD                      & 2048     & -          & -           & 16                  & 23.52             & 39.04              & 34.8     & 0.452\\
PatchGD                 & 2048     & 128        & 10          & 16                  & 26.40             & 32.52              & 53.9     & 0.627\\       
GD-fp16      & 2048     & -          & -           & 24                  & 23.52             & 39.04              & 50.6     & 0.658\\
PatchGD-fp16 & 2048     & 128        & 10          & 24                  & 26.40             & 32.52              & 56.1     & 0.662\\       
GD-fp16                      & 4096     & -          & -           & 24                  & 23.52             & 9.23               & 50.1     & 0.611  \\
PatchGD-fp16                 & 4096     & 128        & 10          & 24                  & 26.41             & 8.09               & 53.5     & 0.667  \\
PatchGD-fp16                 & 4096     & 256        & 10          & 24                  & 26.40             & 9.62               & 55.6     & 0.672  \\ \bottomrule
\end{tabular}
}
\label{tab-panda1}
\end{table*}

\textbf{Prostate Cancer Classification (PANDA). }Table \ref{tab-panda1} presents the results obtained on PANDA dataset for three different image resolutions. For all experiments, we maximize the number of images used per batch while also ensuring that the memory constraint is not violated. For images of $512\times 512$, we see that GD as well as PatchGD deliver approximately similar performance scores (for both accuracy as well as QWK) at 16 GB memory limit. However, for the similar memory constraint, when images of size $2048 \times 2048$ (2K) pixels are used, the performance of GD drops by approximately 10\% while our PatchGD shows a boost of 9\% in accuracy. There are two factors that play role in creating such a big gap in the performance of GD and PatchGD. First, due to significantly increased activation size for higher-resolution images, GD faces the bottleneck of batch size and only 1 image per batch is permitted. Note that to stabilize it, we also experimented with gradient-accumulation across batches, however, it did not help.  Alternatively, we performed hierarchical training, where the model trained on the lower resolution case was used as the initial model for the higher-resolution. To alleviate the issue of using only 1 image per batch, we considered a higher memory limit. Another reason for the low performance is that for higher-resolution images, the optimized receptive field of ResNet50 is not suited which leads to non-optimal performance. 

For increased batch size at 2K resolution, we also considered running quantized networks at half-precision and increased memory (see Table \ref{tab-panda1}). At half-precision, the performance of GD improves, however, it is still significantly lower than PatchGD. Similar observation is made for 4K images that PatchGD performs better. The performance improves further when a patch size of 256 is used. Clearly, from the results reported on PANDA dataset, it is evident that PatchGD is significantly better than GD in terms of accuracy as well as QWK when it comes to handle large images in an end-to-end manner. We also report the latency of both the methods during inference time, and it can be seen that PatchGD performs almost at par with GD. The reason is that unlike GD, the activations produced by PatchGD are smaller and the gain in terms of speed from this aspect balance the slowness induced by patchwise processing of the images. Clearly for applications demanding to handle large images but also aiming to achieve real-time inference, PatchGD could be an interesting direction to explore further.

\textbf{Additional study. }We demonstrated in the earlier experiments that PatchGD performs significantly better than its counterpart. We present here a brief study related to some of the hyperparameters involved in PatchGD. \mbox{Table \ref{tab:sample-ablation}} presents the influence of patch sampling on the overall performance of PatchGD. We vary the sampling fraction per inner-iteration as well as the fraction of samples considered in total for an image in a certain iteration. We observe that keeping the sampling fraction per inner-iteration small helps to achieve better accuracy. This is counter-intuitive since smaller fractions provide a lesser context of the image in one go. We speculate that similar to mini-batch gradient descent, not using too large patch batch size induces regularization noise, which in turn improves the convergence process. However, this aspect needs to be studied in more details for a better understanding. We also observed that the fraction of the image seen in one overall pass of the image in PatchGD does not generally affect the performance, unless it is low. For lower fractions, it is hard for the model to build the global context and the convergene is sub-optimal.

We have also briefly studied the influence of gradient accumulation length parameter for PatchGD and the results are reported in Table \ref{table-epsilon-study} of the appendices. We observed that performing gradient-based model update per inner iteration leads to superior performance for the chosen experiment. However, the choice of $\epsilon$ depends on the number of inner steps $\zeta$. For large values of $\zeta$, values greater than 1 are favored. For example, for the case of processing 2K resolution images with patch size of $128\times 128$, $\epsilon = \zeta$ worked well. However, an empirical relation between $\zeta$ and $\epsilon$ is still to be identified, and this is a part of our future research work.


\begin{table}[]
\footnotesize
\centering
\caption{Sampling ablation on PANDA dataset. Memory limit is 16 GB, Image size and patch size are 2048 and 128 respectively}
\begin{tabular}{@{}ccll@{}}
\toprule
Sampling & Max Sampled  & Accuracy & QWK \\ \midrule
50       & 100     &  42.3        &  0.538   \\
30       & 100     & 49.9         & 0.613    \\
10       & 100          & 53.9       &  0.627   \\
10       & 70           & 53.1         & 0.624     \\
10       & 50           & 53.9       & 0.622  \\
10       & 30           & 51.1         & 0.610   \\ \bottomrule
\end{tabular}
\label{tab:sample-ablation}
\end{table}

\section{Discussion}
\textbf{Hyperparameter optimization and fine-tuning. }PatchGD involves several hyperparameters and their optimized combination is still to be identified. While we have demonstrated the influence through a few experiments, more clarity needs to be gained on the best values of number of inner-update steps to be combined in gradient accumulation ($\epsilon$), striking the right balance between patch size and the number of inner iterations for a given compute memory limit as well choosing the right pretraining strategy. We have observed that using the models trained with GD as the initial models in PatchGD can improve the overall performance. However, there are instances when model training on GD is not possbile. In such scenarios, one could use low-resolution models trained on GD or even the conventional pretrained models. Nevertheless, the effect of each of these choices needs to be thoroughly studied.

\textbf{Application to other tasks. }In this paper, we have focused on demonstrating the working of PatchGD on tasks of image classification, and in particular those where features exist at varying scales. However, this does not limit the applicability of our method to other problems. PatchGD can also be used on the conventional classification problems, and we speculate that it could help to refine the receptive field of the existing models. We discuss this in more details later in this paper. Beyond classification, it is also straightforward to adapt this method for other tasks such as segmentation, object detection, among others, and we intend to cover them in an extended version of this study later.

\textbf{Limitations. }This paper presented the foundational concept of PatchGD. Although we have demonstrated the efficacy of PatchGD through multiple numerical experiments, the overall investigation is still limited in terms of understanding the generalization and stability of the method. Another minor limitation is that since our approach looks only at a fraction of an image in one step, it is relatively slower than the standard gradient descent method. However, since the inference speed is almost the same, this issue creates a bottleneck only when real-time training is a priority.

\textbf{Conclusions. }In this paper, we have demonstrated that it is possible to handle large images with CNN even when the available GPU memory is very limited. We presented Patch Gradient Descent (PatchGD), a novel CNN training strategy that performs model updates  using only fractions of the image at a time while also ensuring that it sees almost the full context over a course of multiple steps. We have demonstrated through multiple experiments the efficacy of PatchGD in handling large images as well as operating under low memory conditions, and in all scenarios, our approach outperforms the standard gradient descent by significant margins. We hope that the details of the method as well as the experimental evidence presented in the paper sufficiently justify the significance of PatchGD in making existing CNN models work with large images without facing the bottleneck of compute memory.

\textbf{Future work. }This paper has established the foundational concept of patch gradient descent to enable training CNNs using very large images. The results as well as insights presented in the paper open doors to several novel secondary research directions that could be interesting in terms of improving the efficacy as well as the acceptance of the presented method in a broader scientific community. Examples of these include extending PatchGD to work on gigapixel images at small compute memory, using PatchGD for enhanced receptive field on standard computer vision tasks, and lastly to couple PatchGD with transformers. Details on the associated challenges and possible modifications are further discussed in Appendix \ref{sec-futute-work}.

\textbf{Acknowledgement}
We would like to thank Texmin Foundation for the financial support provided through grant PSF-IH-1Y-022 to support this work.

\bibliography{example_paper}
\bibliographystyle{icml2022}

\newpage
\appendix
\onecolumn

\section{Future work}
\label{sec-futute-work}
This paper has established the foundational concept of patch gradient descent to enable training CNNs using very large images and even when only limited GPU memory is available for training. The results as well as insights presented in the paper open doors to several novel secondary research directions that could be interesting in terms of improving the efficacy as well as the acceptance of the presented method in a broader scientific community. We list some such directions here.
\vspace{-0.5em}
\begin{itemize}[leftmargin=*]
    \item \textit{Scaling to gigapixel images at small compute memory. }An ambitious but very interesting application of PatchGD would be to be able to process gigapixel images with small GPU memory. We can clearly envision this with PatchGD but with additional work. One important development needed is to extend the PatchGD learning concept to mutiple hierarchical $\mathbf{Z}$ blocks, thereby sampling patches from the outer block to iteratively fill the information in the immediate inner $Z$ block and so on.
    \item \textit{Enhanced receptive field. }So far, PatchGD has been looked at only in the context of being able to handle very large images. However, a different side of its use is that with almost the same architecture, it builds a smaller receptive build, thereby zooming in better. We speculate that in this context, PatchGD could also help in building better discriminative models with lighter CNN architectures. Clearly, this would be of interest to the deep learning community and needs to be explored.
    \item \textit{PatchGD with Transformers. }Transformers are known to provide a better global context and it would be interest to expand the capability of transformers as well to handle large images using PatchGD.
\end{itemize}

\section{Datasets}
\label{app-datasets}
For the purpose of this study, we choose to focus on two cases of large-scale images. We work on different sets of resolutions ranging from 512 to 4096.
\subsection{PANDA}
The Prostate cANcer graDe Assessment Challenge \cite{pandas} consists of one of the largest publically available datasets for Histopathological images which scale to a very high resolution. It is important to mention that we do not make use of any masks as in other aforementioned approaches. Therefore, the complete task boils down to taking an input high-resolution image and then classifying them into 6 categories based on the International Society of Urological Pathology (ISUP) grade groups. The images are downscaled to 4096x4096 resolution applying white bordering to avoid distortion. This is further downscaled to 512x512 and 2048x2048 for demonstrating performance of both the baseline and PatchGD.

\subsection{UltraMNIST}
\label{app-umnist}
This is a synthetic dataset generated by making use of the MNIST digits. For constructing an image, 3-5 digits are sampled such that the total sum of digits is less than 10. Thus an image can be assigned a label corresponding to the sum of the digits contained in the image. Each of the 10 classes from 0-9 each has 1000 samples making the dataset sufficiently large. Note that the variation used in this dataset is an adapted verison of the original data presented in \citet{gupta2022umnist}. Since the digits vary significantly in size and are placed far from each other, this dataset fits well in terms of learning semantic coherence in a image. Moreover, it poses the challenge that downscaling the images leads to significantly loss in the information. While even higher resolution could be chosen, we later demonstrate that the chosen image size is sufficient to demonstrate the superiority of PatchGD over the conventional gradient descent method.
\begin{table}[]
\caption{Influence of different number of gradient accumulation steps $epsilon$ on the performance of MobileNetV2 for the UltraMNIST classification task.}
\centering
\begin{tabular}{@{}lrrcc@{}}
\toprule
Memory & \multicolumn{1}{l}{Image size} & \multicolumn{1}{l}{Patch size} & Grad. accum. iter. ($\epsilon$) & \multicolumn{1}{c}{Accuracy} \\ \midrule
16 GB & 512 & 256 & 1 & 83.7 \\
16 GB & 512 & 256 & 2 & 81.5 \\
16 GB & 512 & 256 & 4 & 81.1 \\ \bottomrule
\end{tabular}
\label{table-epsilon-study}
\end{table}
\begin{figure*}[!ht]
    \centering
    \begin{subfigure}{0.9\linewidth}
    \centering
    \includegraphics[width=\textwidth]{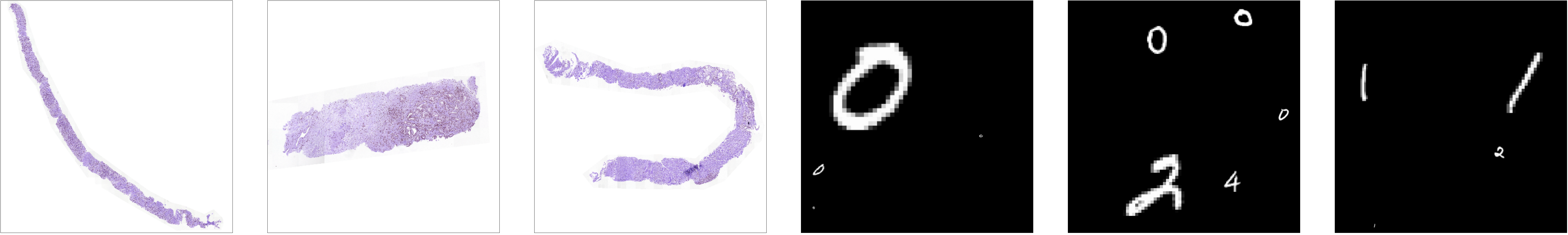}
    \vspace{-1.7em}
    \label{fig-latent}
    \end{subfigure}\vspace{1em}
    \caption{Sample PANDA and UltraMNIST dataset images used for training PatchGD.}
    \label{fig-panda-xai}
\end{figure*}
\label{sec-app}

\section{Training Methodology}
\label{app-train-hyp}
All models are trained for 100 epochs with Adam optimizer and a peak learning rate of 1e-3. A learning rate warm-up of for 2 epochs starting from 0 and linear decay for 98 epochs till half the peak learning rate was employed. The latent classification head consists of 4 convolutional layers with 256 channels in each. We perform gradient accumulation over inner iterations for better convergence, in the case of PANDA. To verify if results are better, not because of an increase in parameters (coming from the classification head), baselines are also extended with a similar head. GD-extended, for MobileNetV2 on UltraMNIST, refers to baseline extended with this head. 

In the case of low memory, as demonstrated in the UltraMNIST experiments, the original backbone architecture is trained separately for 100 epochs. This provides a better initialization for the backbone and is further used in PatchGD. 

For baseline in PANDA at 2048 resolution, another study involved gradient accumulation over images, which was done for the same number of images that can be fed when the percent sampling is 10\%  i.e. 14 times since a 2048x2048 image with a patch size of 128 and percentage sampling of 10 percent can have a maximum batch size of 14 under 16GB memory constraint. That is to say, baseline can virtually process a batch of 14 images. This, however, was not optimal and the peak accuracy reported was in the initial epochs due to the loading of pre-trained model on the lower resolution after which the metrics remained stagnant(accuracy: 32.11\%, QWK:0.357). 

\section{Results: Additional details}
\begin{table}[]
\caption{Influence of different number of gradient accumulation steps $epsilon$ on the performance of MobileNetV2 for the UltraMNIST classification task.}
\centering
\begin{tabular}{@{}lrrcc@{}}
\toprule
Memory & \multicolumn{1}{l}{Image size} & \multicolumn{1}{l}{Patch size} & Grad. accum. iter. ($\epsilon$) & \multicolumn{1}{c}{Accuracy} \\ \midrule
16 GB & 512 & 256 & 1 & 83.7 \\
16 GB & 512 & 256 & 2 & 81.5 \\
16 GB & 512 & 256 & 4 & 81.1 \\ \bottomrule
\end{tabular}
\label{table-epsilon-study}
\end{table}
\begin{figure*}[!ht]
    \centering
    \begin{subfigure}{0.9\linewidth}
    \centering
    \includegraphics[width=\textwidth]{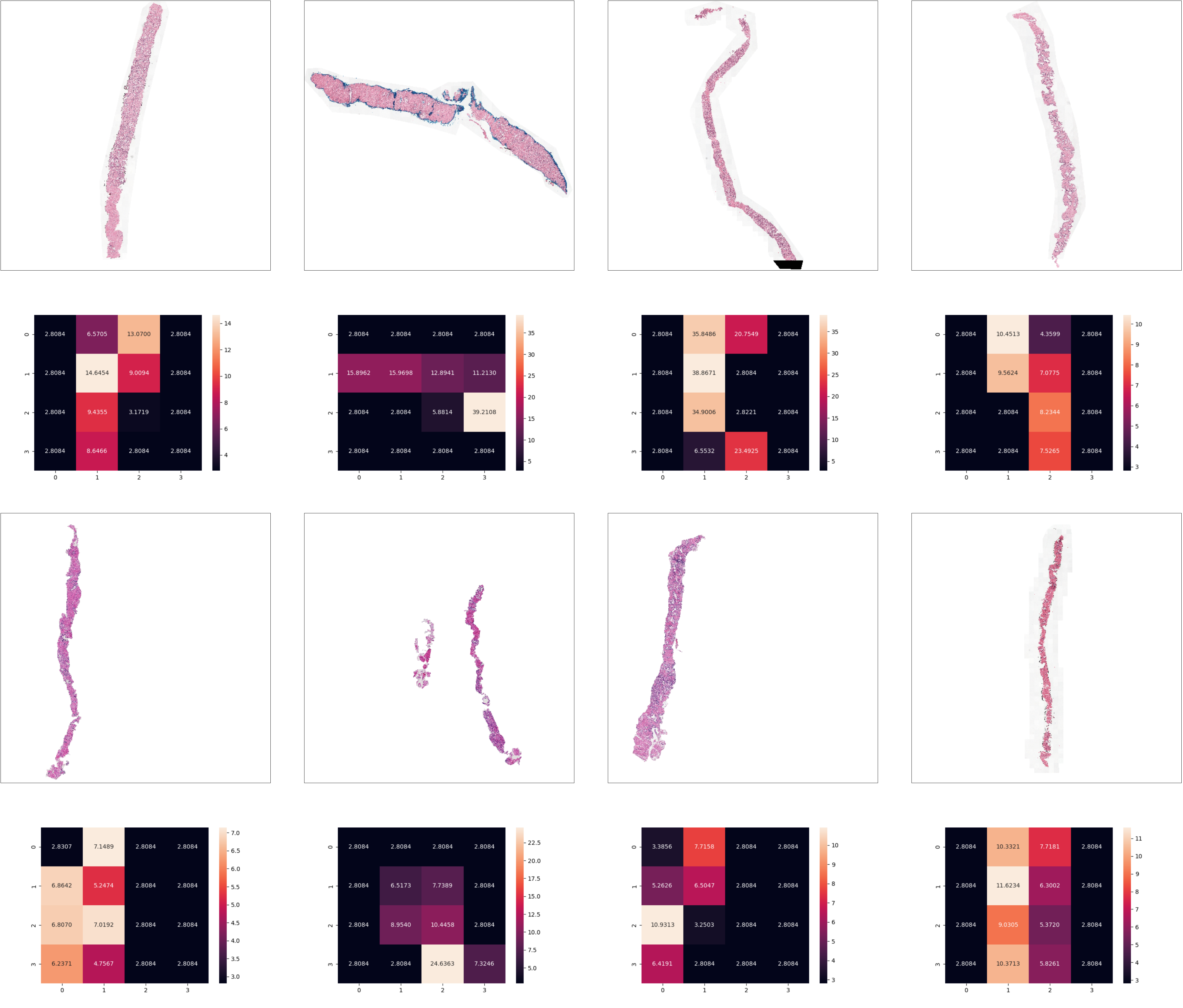}
    \vspace{-1.7em}
    \label{fig-latent}
    \end{subfigure}\vspace{1em}
    \caption{Sample PANDA images along with their latent space $Z$. It can be seen that the latent space clearly acts as a rich feature extractor.}
    \label{fig-panda-xai}
\end{figure*}
\label{sec-app}

\end{document}